\documentclass{article}

\usepackage{microtype}
\usepackage{graphicx}
\usepackage{booktabs} % for professional tables

\usepackage{tabu}
\usepackage{xcolor}
\usepackage{multirow}
\usepackage{subcaption}
\usepackage{caption}
\usepackage{enumitem}

\usepackage{hyperref}

\usepackage[accepted]{icml2023}

\usepackage{amsmath}
\usepackage{amssymb}
\usepackage{mathtools}
\usepackage{amsthm}

\usepackage[capitalize,noabbrev]{cleveref}

\theoremstyle{plain}

\theoremstyle{definition}

\theoremstyle{remark}

\usepackage[textsize=tiny]{todonotes}

\icmltitlerunning{Membrane Potential Distribution Adjustment and Parametric Surrogate Gradient in SNN}

\begin{document}

\twocolumn[
\icmltitle{Membrane Potential Distribution Adjustment and Parametric Surrogate Gradient in Spiking Neural Networks}

% It is OKAY to include author information, even for blind
% submissions: the style file will automatically remove it for you
% unless you've provided the [accepted] option to the icml2023
% package.

% List of affiliations: The first argument should be a (short)
% identifier you will use later to specify author affiliations
% Academic affiliations should list Department, University, City, Region, Country
% Industry affiliations should list Company, City, Region, Country

% You can specify symbols, otherwise they are numbered in order.
% Ideally, you should not use this facility. Affiliations will be numbered
% in order of appearance and this is the preferred way.
\icmlsetsymbol{equal}{*}

\begin{icmlauthorlist}
\icmlauthor{Siqi Wang}{yyy}
\icmlauthor{Tee Hiang Cheng}{yyy}
\icmlauthor{Meng-Hiot Lim}{yyy}
% \icmlauthor{Firstname4 Lastname4}{sch}
% \icmlauthor{Firstname5 Lastname5}{yyy}
% \icmlauthor{Firstname6 Lastname6}{sch,yyy,comp}
% \icmlauthor{Firstname7 Lastname7}{comp}
%\icmlauthor{}{sch}
% \icmlauthor{Firstname8 Lastname8}{sch}
% \icmlauthor{Firstname8 Lastname8}{yyy,comp}
%\icmlauthor{}{sch}
%\icmlauthor{}{sch}
\end{icmlauthorlist}

\icmlaffiliation{yyy}{School of Electrical and Electronic Engineering, Nanyang Technological University, Singapore}
% \icmlaffiliation{comp}{Company Name, Location, Country}
% \icmlaffiliation{sch}{School of ZZZ, Institute of WWW, Location, Country}

\icmlcorrespondingauthor{Siqi Wang}{siqi002@e.ntu.edu.sg}
% \icmlcorrespondingauthor{Firstname2 Lastname2}{first2.last2@www.uk}

% You may provide any keywords that you
% find helpful for describing your paper; these are used to populate
% the "keywords" metadata in the PDF but will not be shown in the document
\icmlkeywords{Spiking neural networks (SNN), Potential distribution, Surrogate gradient, SNN learning enhancement}

\vskip 0.3in
]

% this must go after the closing bracket ] following \twocolumn[ ...

% This command actually creates the footnote in the first column
% listing the affiliations and the copyright notice.
% The command takes one argument, which is text to display at the start of the footnote.
% The \icmlEqualContribution command is standard text for equal contribution.
% Remove it (just {}) if you do not need this facility.

\printAffiliationsAndNotice{}  % leave blank if no need to mention equal contribution
% \printAffiliationsAndNotice{\icmlEqualContribution} % otherwise use the standard text.

\begin{abstract}
As an emerging network model, spiking neural networks (SNNs) have aroused significant research attentions in recent years. However, the energy-efficient binary spikes do not augur well with gradient descent-based training approaches. Surrogate gradient (SG) strategy is investigated and applied to circumvent this issue and train SNNs from scratch. Due to the lack of well-recognized SG selection rule, most SGs are chosen intuitively. We propose the parametric surrogate gradient (PSG) method to iteratively update SG and eventually determine an optimal surrogate gradient parameter, which calibrates the shape of candidate SGs. In SNNs, neural potential distribution tends to deviate unpredictably due to quantization error. We evaluate such potential shift and propose methodology for potential distribution adjustment (PDA) to minimize the loss of undesired pre-activations. Experimental results demonstrate that the proposed methods can be readily integrated with backpropagation through time (BPTT) algorithm and help modulated SNNs to achieve state-of-the-art performance on both static and dynamic dataset with fewer timesteps.
\end{abstract}

\section{Introduction}
In recent years, spiking neural networks (SNNs), have commanded increasing attentions from researchers due to their event-driven sparsity, high computational efficiency, and accurate mimicry of human brains. Benefiting from utilizing binary spike-based states and characterizing temporal effect among neurons, SNNs explicitly take advantage of sparse spatiotemporal information processing, and thus, dubbed as the third generation of neural networks. These properties permit SNNs to be energy efficient, making it suitable for real-time applications on power-restricted edge devices. Moreover, neuromorphic hardware has been progressively developed to be compatible with the dynamics of spiking neural mechanisms to accelerate SNN models construction.

However, the intrinsic step activation function of spiking neurons prohibits synaptic weights update through gradient descent due to its non-differentiability. To circumvent this problem and effectively train SNNs, two main methods have proven to be efficient: ANN-to-SNN conversion and gradient descent approach based on surrogate gradient (SG). The conversion method \cite{cao2015spiking,sengupta2019going,yu2021constructing} obtains SNN synaptic weights based on a pre-trained ANN with the same structure, through transferring neurons’ activations in ANN to the corresponding firing rates. Although the converted SNNs possess the capability to achieve nearly lossless results as the original ANNs, there is high latency in ensuring that ANN neurons’ activation values are accurately represented by spiking neurons’ firing rates. On the other hand, SNNs constructed with SG method \cite{wu2018spatio,cheng2020lisnn,perez2021sparse} utilize auxiliary functions to replace the non-differentiable neural activation function to iteratively minimize network loss and update their connections with backpropagation through time (BPTT) scheme \cite{thiele2019spikegrad}. Adopting BPTT framework, network loss can be backpropagated through both layer-wise spatial domain and timestep-wise temporal domain. SG method provides an opportunity to directly train SNNs within several timesteps, making it possible to process real-time data. Unfortunately, as with normal gradient descent training strategy, SG approaches also face the issue of gradient vanishing problems. Besides, quantization errors caused by improper potential distributions will be accumulated when network becomes deeper. Various techniques have been initiated or leveraged from ANNs to combat performance degradation in SNNs. However, analysis of potential distributions variation and quantization error encountered when generating spikes is usually not involved in those techniques.

As more low-latency and energy-saving learning algorithms are being developed, diverse auxiliary functions and corresponding SGs have been chosen to formulate different learning strategies \cite{wang2022hierarchical}. Researchers usually select SGs for their models based on experience and preferences. Commonly used SGs contain different graphical shapes such as triangular function, rectangular function, arctan function, \textit{etc}. For each candidate SG, parameters that control its value range or curvature, all are empirically defined and remain constant throughout training due to the lack of widely recognized SG selection rule. However, the chosen SG and parameters may not always be the optimal ones for the target model or task, leaving room for doubts on the effects of different SGs on model performance. The suboptimal SGs may decelerate network convergence and lead to convexity issue.

In this paper, we incorporate a potential distribution adjustment method and a parametric surrogate gradient mechanism to address these problems. The main contribution of this work can be summarized as follows:

\vspace{-\topsep}
\begin{itemize}[leftmargin=1.3em]
    \itemsep0em
    \item [1)]
    We propose membrane potential distribution adjustment (PDA) method by incorporating potential distribution loss ($\mathcal{L}_{PD}$) to quantify the inconsistency between actual neural potential distribution and target. The $\mathcal{L}_{PD}$ can be integrated into the overall network loss so that membrane potentials in each layer can be redistributed by synaptic weights update with network convergence. The potential distribution loss minimization leads to data normalization, quantization error reduction, and capability improvement on deep SNN construction.
    \item [2)]
    We evaluate the results of utilizing different SGs for a specific classification task, quantifying the effect of different surrogate gradient parameter selections. A new parametric surrogate gradient (PSG) strategy is proposed to parameterize the potential SG and to find the optimal parameter values through training. Issue of suboptimal SG can be solved by using the PSG method.
    \item [3)]
    To evaluate the effectiveness of the proposed methods, we implement them on ResNet architectural network and train the SNN for both static (MNIST, CIFAR10) and neuromorphic (NMNIST, DVS-CIFAR10) datasets. To the best of our knowledge, our SNNs produced better results in almost all instances of datasets tested compared to other state-of-the-art works in terms of accuracies, clearly distinguishing the beneficial effects and its extraordinary performance enhancement on SNNs.
\end{itemize}
\vspace{-\topsep}

\section{Background and Related Work}
\subsection{Leaky Integrate-and-Fire Mechanism}\label{section2A}
A spiking neuron’s state represents its output to the following layers in every timestep in SNNs. There are only two possible neural states for a spiking neuron, either fire (output value “1”) or stay silent (output value “0”). Spikes will be generated if neuron’s membrane potential reaches the threshold ($V_{th}$), otherwise, the neuron remains inactive.

\begin{figure}
% \vskip 0.1in
  \centering
  \includegraphics[width=\columnwidth]{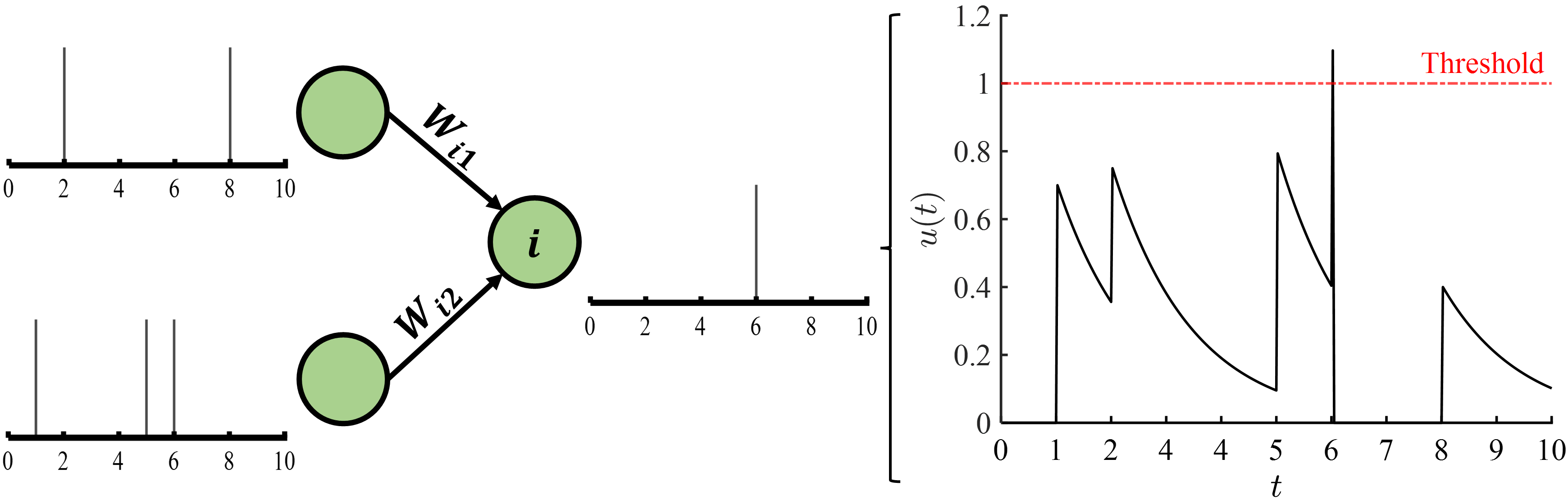}
\caption{Illustration of postsynaptic neuron membrane potential update using LIF model.}
\label{postsynaptic_response}
% \vskip -0.1in
\end{figure}

To modulate neuron’s behaviors and simulate on computers, various approximated neural mechanisms have been introduced by neuroscientists. Trade-off between the model’s computational complexity and biological plausibility \cite{izhikevich2004model} always exists due to the complex and delicate dynamics in neuro-biological brain systems. Among the adopted mechanisms, the Leaky Integrate-and-Fire (LIF) is commonly recognized to be suitable for spiking neuron simulation. Neural behaviors such as potential accumulation, spike firing, and state reset characterized by LIF can be depicted as
\begin{equation}\label{eq2}
    \tau\frac{du(t)}{dt}=-u(t)+I(t)
\end{equation}
In this work, we chose LIF as the neural model due to its simple computation and characteristically sufficient biological plausibility. Both the initial state and reset potential are set to zero for all neurons in the following experiments. Fig.~\ref{postsynaptic_response} illustrates how a neuron responses to input spike trains.

Eq.~\ref{eq2} demonstrates the continuous state of a spiking neuron. However, neural dynamics in discrete timesteps are usually more suitable for simulations. There are two common ways to discretize and calculate a neuron’s membrane potential at specific timestep. One could discretize Eq.~\ref{eq2} first and then calculate the potential or alternatively, derive continuous potential expression first and then discretize it. We use the latter in this work as less errors will be incurred because discretization only happens in the last derivation step. The membrane potential and neuron’s state can be derived as follow:
\begin{equation}\label{eq3}
    u_{i}^{t,n}=\varepsilon u_{i}^{t-1,n}(1-o_{i}^{t-1,n})+\sum_{j\in l(n-1)}W_{ij}o_{j}^{t,n-1}
\end{equation}
\begin{equation}\label{eq4}
    o_{i}^{t,n}=f(u_{i}^{t,n})
\end{equation}
where $\varepsilon=e^{\frac{1}{\tau}}$ is the decay factor, the superscript $t$ and $n$ represent timestep and layer index, respectively. $f(\cdot)$ in Eq.~\ref{eq4} is a step activation to determine neuron’s output based on internal potential.

\subsection{Quantization Error}
Due to the binary transmission medium in SNNs, membrane potential distribution will shift unpredictably as data pass through multiple layers \cite{guo2022recdis}. Neurons’ potential values may fall into inappropriate parameter update region of the SG and finally degrade the training process. Step activation used in SNNs will assign binary values as output according to the corresponding neuronal states. In this case, exact potential values are not essential and gaps between threshold and actual potentials will be ignored. Two neurons will be indistinguishable as long as their outputs are the same. Such indistinguishability is classified as quantization error.

Many SNN oriented normalization techniques have been presented to redistribute pre-activations in a manner such that proper number of neurons would associate with the selected SG and update accordingly. The NeuNorm \cite{wu2019direct} can harmonize neural selectivity and normalize neurons’ activities. Furthermore, the threshold dependent batch normalization \cite{zheng2021going} realizes reallocating potentials in SNNs through normal distribution in order to balance active neurons, hence enhances performance.

\subsection{Regularization}
Since most current SNNs intend to have deeper architectures to achieve higher nonlinearities for better performance, the risk of overfitting in the trained model also becomes unprecedentedly higher. Dropout is one of the most typical techniques to prevent overfitting in deep network construction and has been successfully leveraged and verified to be beneficial in SNN \cite{lee2020enabling,kappel2015network}. Compared with the original version for ANNs, dropping probabilities used in SNNs are usually of smaller values as events flowing inside SNNs are sparser. Moreover, various dropout variants have been proposed to further improve SNNs’ performance. For example, the moderate dropout \cite{wangltmd} is devised to not only exclude overfitting but also stabilize the modulated SNN by minimizing inconsistences between sub-networks generated by dropout. In this work, we adopt the moderate dropout in our SNNs and set the dropping probability to be 0.2 so that sufficient neurons remain for data transmission to be feasible.

\subsection{Extension of Parametric Neural Mechanism}
Spiking neuronal mechanism plays an essential role in deciding neuronal behaviors under external stimulus, thus have received special attentions from researchers. Multiple extensions of parametric neural mechanism have been developed to enhance biological plausibility and model performance in recent years. Lateral interactions among neighboring neurons are adopted to enhance model robustness to noise \cite{cheng2020lisnn}. Learnable time constant is introduced to achieve better accuracy and more feasible neuronal behaviors in \cite{zimmer2019technical} and \cite{fang2021incorporating} for speech recognition and image classification tasks. In \cite{wangltmd}, learnable thresholding mechanism is designed to enhance network learning capability and to balance the number of firing neurons in SNNs. Recognizing on the beneficial effects of these techniques, in this work, our models are constructed with lateral interactions and learnable thresholding neurons. However, all the above approaches focus on neuronal features themselves, but omit the influence of the activation function that actually determines neurons’ behaviors, which is exactly the aim of the proposed PSG method in this paper.

\section{Methods}

\subsection{Membrane Potential Distribution Adjustment}\label{section3A}
A postsynaptic neuron’s behavior depends on both its previous state and input from preceding layers as denoted in Eq.~\ref{eq3}. In this scenario, binarized output from presynaptic neurons will lead to fluctuating postsynaptic neuronal states and shifted membrane potentials distributions. Undesired potential distribution would slow down training process or even mislead network convergence direction since neurons’ rectifications are controlled by their SGs at corresponding potential values. To circumvent this, we propose the PDA method by defining a potential distribution loss ($\mathcal{L}_{PD}$) to embed potential distribution punishment into the overall network objective function so that potentials can be updated automatically as the network converges.

In order to measure the loss of potential distribution, a target pre-activation distribution is necessary. It has been clarified in \cite{chen2021pruning} that weight distribution in direct trained deep SNNs is normally unimodal so that neurons’ potentials should be analytically a Gaussian distribution for each layer, labeled as $u\in \mathcal{N}(\mu,\sigma^{2})$. Consequently, we define the difference between the actual potential distribution and the standard Gaussian distribution as $\mathcal{L}_{PD}$ and consider it as part of the network loss to be minimized during training by gradient descent. In this work, we use the Kullback–Leibler (KL) divergence to statistically measure the distance between the two distributions as shown in Eq.~\ref{eq5}.
\begin{equation}\label{eq5}
    D_{KL}(P^{t,n}\parallel \mathcal{N}^{n})=\sum_{x}P^{t,n}(x)\log \bigg( \frac{P^{t,n}(x)}{\mathcal{N}^{n}(x)} \bigg)
\end{equation}
where $P^{t,n}$ and $\mathcal{N}^{n}$ are the actual and standard Gaussian potential distributions in layer $n$ at time $t$, respectively. It is noted that the KL divergence depicted in Eq.~\ref{eq5} is asymmetric, such that $D_{KL}(P^{t,n}\parallel \mathcal{N}^{n})\neq D_{KL}(\mathcal{N}^{n}\parallel P^{t,n})$. In this respect, we involve both measurements in our loss equation to make it symmetric. Induced symmetry would enhance robustness by preventing the very small valued asymmetric KL divergence from corroding the impact of the overall loss. Furthermore, an additional coefficient $\beta$ is introduced to adjust the impact of potential distribution loss in the overall network loss function as follows:
\begin{gather}
\begin{aligned}
    \mathcal{L}_{PD}=\frac{1}{2TN}\sum_{t=1}^{T}\sum_{n=1}^{N}\big(D_{KL}&(P^{t,n}\parallel \mathcal{N}^{n}) \\
    &+D_{KL}(\mathcal{N}^{n}\parallel P^{t,n})\big)\label{eq6}
\end{aligned} \\
    \mathcal{L}=Loss(\frac{1}{T}\sum_{t=1}^{T}o^{t,N},Y)+\beta\mathcal{L}_{PD}\label{eq7}
\end{gather}
where $Y$ is the target output and $Loss$ represents the common loss function used in SNN models. The overall loss $\mathcal{L}$ will be used to update all the learnable parameters according to the gradient descent scheme. Coefficient $\beta$ is set to be 1 for all the experiments in this paper. Eq.~\ref{eq7} ensures that pre-activation distribution of each layer is incorporated into the overall loss function, leading to more suitable neuronal potential distributions and faster convergence.

\subsection{Parametric Surrogate Gradient}
\begin{figure}
% \vskip 0.1in
  \centering
     \begin{subfigure}{0.49\columnwidth}
         \centering
         \includegraphics[width=\columnwidth]{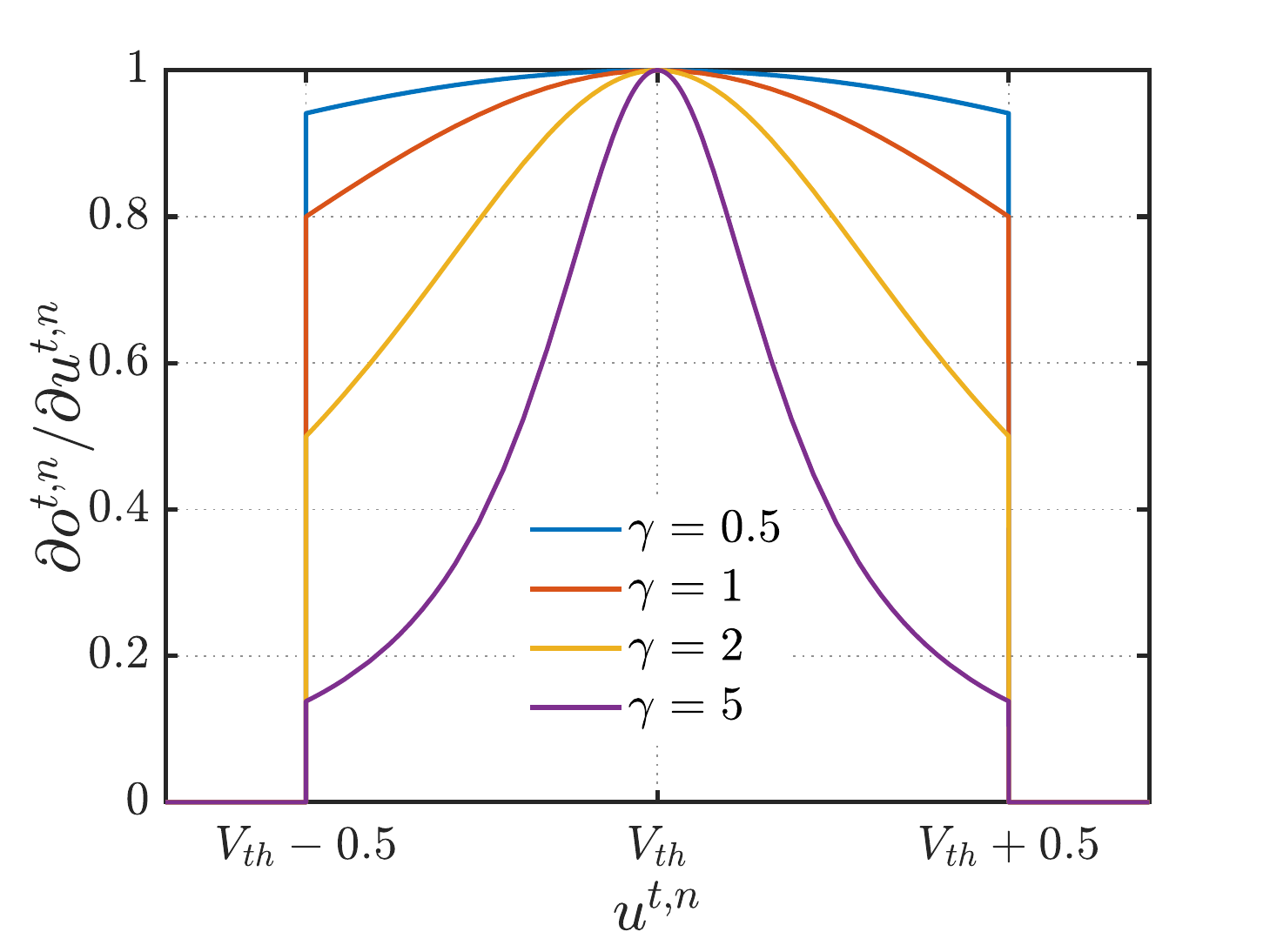}
         \caption{}
         \label{gradient_plot}
     \end{subfigure}
     \hfill
     \begin{subfigure}{0.49\columnwidth}
         \centering
         \includegraphics[width=\columnwidth]{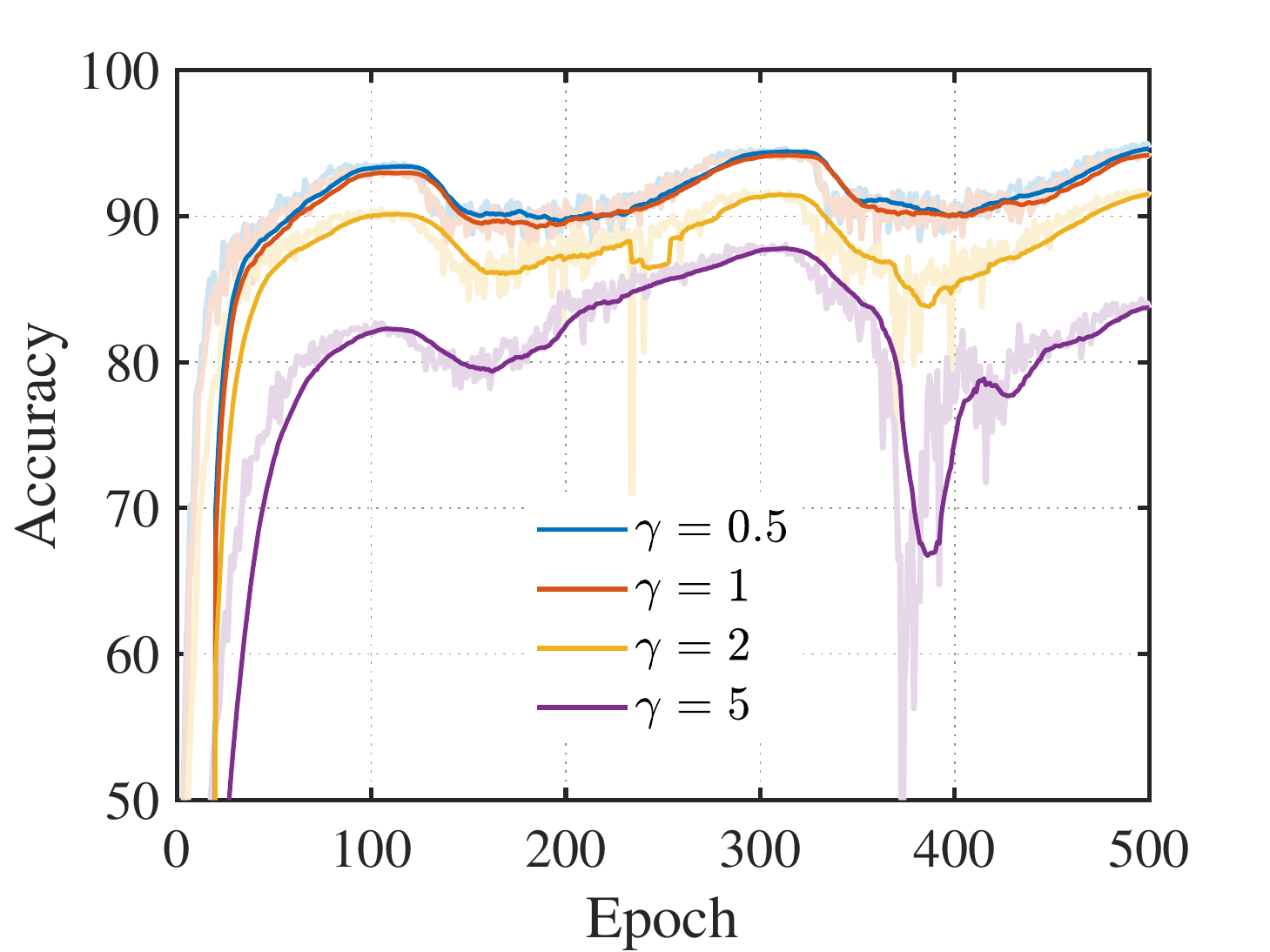}
         \caption{}
         \label{different_SG_result}
     \end{subfigure}
\caption{(a) Different values of surrogate gradient parameter $\gamma$ can result in distinctive surrogate gradients. (b) Inference results of the same SNN configuration but different surrogate gradients. Original data are represented by light-color curves and dark-color curves record moving average of 20~epoches.}
% \vskip -0.1in
\end{figure}

As mentioned in Sections~\ref{section2A} and \ref{section3A}, we adopt LIF model to characterize neurons’ responses to external stimulus and take BPTT as the training framework in our SNNs. Employing BPTT, network error will be backpropagated in both spatial and temporal domains to update network parameters iteratively. The loss function derivative for synaptic weights update required by gradient descent algorithm can be formulated as Eq.~\ref{eq8} as follows:
\begin{equation}\label{eq8}
    \frac{\partial \mathcal{L}}{\partial W^{n}}=\sum_{t=1}^{T}\frac{\partial \mathcal{L}}{\partial o^{t,n}}\frac{\partial o^{t,n}}{\partial u^{t,n}}\frac{\partial u^{t,n}}{\partial W^{n}}
\end{equation}
where the terms $\frac{\partial \mathcal{L}}{\partial o^{t,n}}$ and $\frac{\partial u^{t,n}}{\partial W^{n}}$ can be directly calculated according to the network loss function Eq.~\ref{eq7} and potential update equation Eq.~\ref{eq3}, and $\frac{\partial o^{t,n}}{\partial u^{t,n}}$ is the partial derivative of a neuron’s output to its membrane potential, which is the derivative of the step activation function $f'(u)$ in Eq.~\ref{eq4} for spiking neurons. However, step function is non-differentiable thus gradient descent training algorithm is intrinsically not applicable for SNNs. Auxiliary functions in various forms have been applied to virtually replace the step function so that SGs can be derived for $f'(u)$ calculation to deal with the problem. However, there is no widely recognized criteria for SGs selection in SNNs. Researchers empirically chose the SG for their models for most of the cases. For example, arctangent function $\frac{1}{\gamma}\arctan{(\gamma(x-V_{th}))+\frac{1}{2}}$ is one of the most commonly used auxiliary functions in SNNs, whose gradient is in the form of $\frac{1}{1+(\gamma(x-V_{th}))^{2}}$. The shape of such gradient can be manipulated by changing the value of gradient parameter $\gamma$ as depicted in Fig.~\ref{gradient_plot}.

For most of the current well-behaved SNNs, once the auxiliary function is defined, its gradient will be finalized and is never changed during the network training process. We construct SNNs using different SGs to evaluate their performances on classifying CIFAR10 images with two timesteps. As illustrated in Fig.~\ref{different_SG_result}, we can observe that different SG selections corresponding to those shown in Fig.~\ref{gradient_plot}, can efficaciously affect the SNNs’ inference results on the same task. In other words, an improper SG will limit the performance of SNNs and lead to suboptimal SNN models after training. It validates the importance of SG selection for the final SNN models performance. Unfortunately, only few works pay attention to the influence of surrogate functions selection so far. In \cite{li2021differentiable}, a set of SG candidates is compared to a finite difference gradient (FDG), which is considered as an approximation of the true gradient, and the best SG is finalized by selecting the gradient with the highest similarity. Once the best SG among the candidates is found, SNN will be built accordingly. However, the FDG selection is empirical, and the propriety cannot be guaranteed. In addition, FDG is determined by evaluating the result only in the first layer due to the time-consuming double computation of loss for each weight element. It limits the selected FDG's suitability when implemented for the entire SNN. Despite the existence of several optimal gradient selection methods, a systematic and computationally affordable approach for finding the optimal SG for any particular SNN model still remains as an open question, which is exactly the aim of this paper.

\begin{figure}
% \vskip 0.1in
  \centering
  \includegraphics[width=0.62\columnwidth]{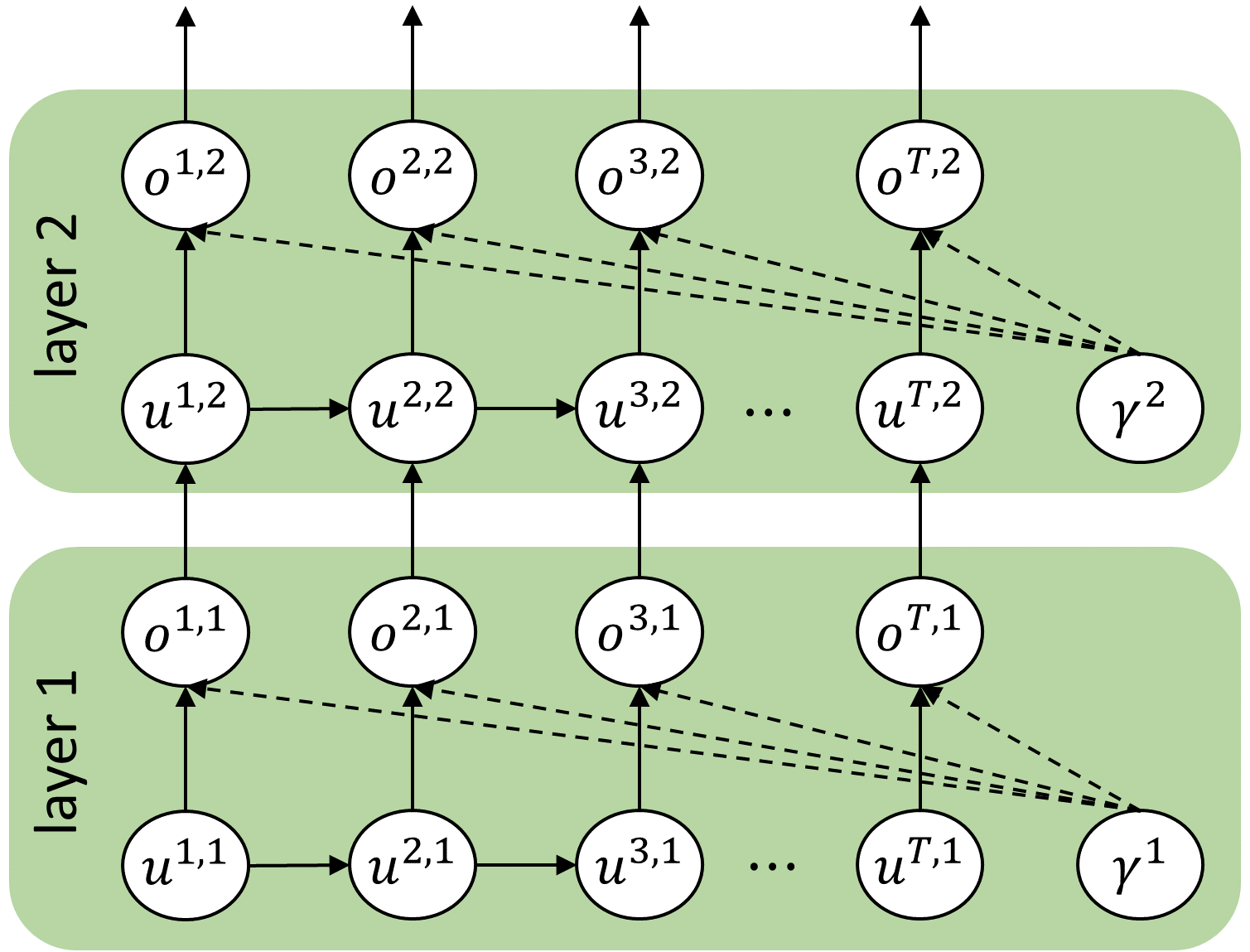}
\caption{Illustration of the PSG in the forward path. Dashed line represents that the surrogate gradient parameter $\gamma$ only participants data backpropagation but not forward direction.}
\label{PSG_flowchart}
% \vskip -0.1in
\end{figure}

To identify the best gradient shapes and fully exploit potential of our SNN models, we propose a systematic approach by defining the form of auxiliary functions for different types of layers and designing a gradient descent-based method to iteratively update surrogate gradient parameter to carve out the gradient shape during network training. With the proposed gradient optimization method, the shapes of SGs are parameterized and able to automatically reach their optima as network converges, which provides a feasible solution to ascertain the most suitable SGs for the current task. To this end, in our proposed PSG method, a suitable auxiliary function must fulfill the following three requirements:
\vspace{-\topsep}
\begin{itemize}[leftmargin=1.3em]
    \itemsep0em
    \item [1)]
    The auxiliary function should be a function of two variables to virtually describe the activation of spiking neurons, formulated as $o=f(u,\gamma)$, where $u$ is the current membrane potential and $\gamma$ is a surrogate gradient parameter to control the SG function shape, as shown in Fig.~\ref{PSG_flowchart}. It should be noted that the defined auxiliary function will only be used to derive the surrogate gradient, but never be involved in data forward propagation. Particularly, step function is consistently used to determine the firing states of all neurons based on their membrane potentials and thresholds. Therefore, in the flowchart, we use dashed lines to represent such non-participation of surrogate gradient parameter $\gamma$ in the forward path.
    \item [2)]
    The auxiliary function must be partially differentiable for all possible potential $u$ and surrogate gradient parameter $\gamma$ values. The partial derivatives $\nabla f(u, \gamma)=[\frac{\partial f}{\partial u},\frac{\partial f}{\partial \gamma}]$ are treated as SGs and will participant in the process of updating synaptic connections and surrogate gradient parameters according to BPTT training algorithm. The values of SGs should be located within the proper value ranges to avoid any extreme high or low gradient value, which may lead to unexpected gradient exploding or vanishing in deep SNNs. 
    \item [3)]
    Because neurons use either firing states or potentials as their output, neuronal potentials distribute inconsistently at different layers of our SNNs. Hence, distinctive auxiliary functions should be utilized for the layers accordingly. For convolutional layers, since only neurons with potential values near the threshold can be updated, $\frac{\partial f}{\partial u}$ is supposed to reach the highest value when $u$ approaches the threshold. While for fully-connected layers, because neurons’ outputs are proportional to their accumulated potential values over time, $\frac{\partial f}{\partial u}$ should have higher values as $u$ becomes large. 
\end{itemize}
\vspace{-\topsep}
According to the above requirements, we separately define the auxiliary functions for convolutional and fully-connected layers as following forms:

\textbf{Case 1: Convolutional layers with spikes output.}
\newline In this case, we adopt the commonly used arctangent function as the basic building block of our auxiliary function, with the parameter $\gamma$ to calibrate the derived SG. 
\begin{equation}\label{eq9}
    f(u,\gamma)=\arctan{[\gamma (u-V_{th})]}+\frac{1}{2}
\end{equation}
Thus, partial derivatives $\frac{\partial f}{\partial u}$ and $\frac{\partial f}{\partial \gamma}$ can be directly obtained from Eq.~\ref{eq9}.
\begin{gather}
    \frac{\partial f}{\partial u}=\frac{\gamma}{1+\bigl(\gamma(u-V_{th})\bigl)^{2}}\label{eq10} \\
    \frac{\partial f}{\partial \gamma}=\frac{u-V_{th}}{1+\bigl((u-V_{th})\gamma\bigl)^{2}}\label{eq11}
\end{gather}
Although the ranges of the values for the above derivatives are restricted by the values of variables $\gamma$ and $u$ in each iteration, it is possible for the derivatives to be trapped in extreme values during training since these variables are not bounded. Consequently, we rescale these derivatives and make the maximum values to be 1 to avoid such unexpected situations. Thus, the PSGs for convolutional layers can be expressed as:
\begin{gather}
    \frac{\partial o}{\partial u}=\frac{\partial f}{\partial u}=\frac{1}{1+\bigl(\gamma(u-V_{th})\bigl)^{2}} \label{eq12}\\
    \frac{\partial o}{\partial \gamma}=\frac{\partial f}{\partial \gamma}=\frac{1}{1+\bigl((u-V_{th})\gamma\bigl)^{2}} \label{eq13}
\end{gather}
\textbf{Case 2: Fully-connected layers with potentials output.}
\newline In this case, we expect the SGs to have relatively high values when the variables become large since neurons in fully-connected layers use accumulated potentials as output, which will generally have large values over time. Auxiliary function defined for convolutional layers is no longer suitable since it vanishes as value of the variable goes to infinity. To this end, a new auxiliary function expression is specifically defined for fully-connected layers as follows:
\begin{equation}\label{eq14}
    f(u,\gamma)=\ln(1+e^{\gamma (u-V_{th})})
\end{equation}
The partial derivatives of Eq.~\ref{eq14}, written as Eq.~\ref{eq15} and \ref{eq16} have larger values when neuronal potentials become high, to match the generally high neuronal potentials in the fully-connected layers.
\begin{gather}
    \frac{\partial f}{\partial u}=\frac{\gamma}{1+e^{-\gamma(u-V_{th})}} \label{eq15}\\
    \frac{\partial f}{\partial \gamma}=\frac{u-V_{th}}{1+e^{-(u-V_{th})\gamma}} \label{eq16}
\end{gather}
Same as convolutional layers, we rescale the above derivatives to restrict their value ranges. The normalized PSGs for fully-connected layers can be derived as follow:
\begin{gather}
    \frac{\partial o}{\partial u}=\frac{\partial f}{\partial u}=\frac{1}{1+e^{-\gamma(u-V_{th})}} \label{eq17}\\
    \frac{\partial o}{\partial \gamma}=\frac{\partial f}{\partial \gamma}=\frac{1}{1+e^{-(u-V_{th})\gamma}} \label{eq18}
\end{gather}
Based on the above two cases, derivative of neuronal output with respect to their potential is defined so that loss function derivative for synaptic connections can be successfully calculated according to Eq.~\ref{eq8}. Furthermore, by using the same principle and illustrated in Eq.~\ref{eq19}, procedure of proper SG selection has been successfully merged into network convergence process by iteratively updating surrogate gradient parameter $\gamma$ via backpropagation.
\begin{equation}\label{eq19}
    \frac{\partial \mathcal{L}}{\partial \gamma^{n}}=\sum_{t=1}^{T}\frac{\partial \mathcal{L}}{\partial o^{t,n}}\frac{\partial o^{t,n}}{\partial \gamma^{n}}
\end{equation}
Therefore, both weights and SG can be iteratively updated. And the gradients for synaptic weights and surrogate gradient parameter using BPTT in convolutional and fully-connected layers can be calculated using the corresponding variable values from the previous iteration.

\textbf{Gradient for synaptic weights.}
\begin{equation}\label{eq20}
  \frac{\partial \mathcal{L}}{\partial W^{n}} =
    \begin{cases}
      \sum_{t=1}^{T}\frac{\partial \mathcal{L}}{\partial o^{t,n}}\frac{1}{1+(\gamma(u-V_{th}))^{2}}\frac{\partial u^{t,n}}{\partial W^{n}}, & \text{Conv layers}\\
      \sum_{t=1}^{T}\frac{\partial \mathcal{L}}{\partial o^{t,n}}\frac{1}{1+e^{-\gamma(u-V_{th})}}\frac{\partial u^{t,n}}{\partial W^{n}}, & \text{FC layers}
    \end{cases}       
\end{equation}
\textbf{Gradient for surrogate gradient parameter.}
\begin{equation}\label{eq21}
  \frac{\partial \mathcal{L}}{\partial \gamma^{n}} =
    \begin{cases}
      \sum_{t=1}^{T}\frac{\partial \mathcal{L}}{\partial o^{t,n}}\frac{1}{1+((u-V_{th})\gamma)^{2}}, & \text{Conv layers}\\
      \sum_{t=1}^{T}\frac{\partial \mathcal{L}}{\partial o^{t,n}}\frac{1}{1+e^{-(u-V_{th})\gamma}}, & \text{FC layers}
    \end{cases}       
\end{equation}

\section{Empirical Evaluation}
We evaluate the proposed PDA and PSG methods by testing the constructed SNNs with classification tasks on both static (MNIST, CIFAR10) and neuromorphic (NMNIST, DVS-CIFAR10) datasets. 

Inference results and simulation timesteps of our SNNs are illustrated in Tab.~\ref{infer_result}, showing that models with the proposed methods outperform other state-of-the-art SNNs even with fewer timesteps on almost all datasets. Since all spatiotemporal connections need to be stored for future neuronal states computation when BPTT algorithm is applied, larger memory space will be required if more timesteps are involved in the constructed SNNs. From the comparison in Tab.~\ref{infer_result}, our SNNs generally require much fewer timesteps than other works, leading to significant improvement on both memory space occupancy and system latency.
\begin{table*}[t]
\vskip -0.1in
\caption{Comparison among state-of-the-art SNNs and the proposed methods on different datasets. Numbers highlighted in italics (in parentheses) and bold represent timesteps required and highest accuracies achieved.}\label{infer_result}
\vskip 0.1in
\centering
\resizebox{\textwidth}{!}{
\begin{small}
\begin{tabular}{ccccccc}
\hline
\textbf{Model} & \textbf{Method} & \textbf{MNIST} & \textbf{CIFAR10} & \textbf{CIFAR100} & \multicolumn{1}{c}{\textbf{NMNIST}} & \multicolumn{1}{c}{\textbf{DVS-CIFAR10}} \\ \hline
\cite{hunsberger2015spiking} & ANN-to-SNN & 98.37\% & 82.95\% &  & - & - \\
% \cite{neil2016effective} & ANN-to-SNN & - & - & - & 95.72\% & - \\
\cite{esser2016convolutional} & ANN-to-SNN & - & 89.32\% & 65.48\% & - & - \\
\cite{diehl2016conversion} & ANN-to-SNN & 99.10\% & - & - & - & - \\
\cite{rueckauer2017conversion} & ANN-to-SNN & 99.44\% & 88.82\% & - & - & - \\
\cite{sengupta2019going} & ANN-to-SNN & - & 91.55\% & - & - & - \\
\cite{han2020rmp} & ANN-to-SNN & - & 93.63\% & 70.93\% & - & - \\
% \cite{cohen2016skimming} & SKIM & - & - & - & 92.87\% (\textit{360}) & - \\
\cite{sironi2018hats} & HATS & - & - & - & - & 52.40\% \\
\cite{ramesh2019dart} & DART & - & - & - & - & 65.78\% \\
\cite{kugele2020efficient} & Streaming rollout ANN & - & - & - & - & 66.75\% \\
\cite{lee2016training} & Direct training & - & - & - & 98.74\% (\textit{300}) & - \\
\cite{lee2018training} & Hybrid direct training &99.28\% (\textit{175}) & - & - & - & - \\
\cite{wu2018spatio} & Direct training & 99.42\% (\textit{30}) & 50.70\% (\textit{30}) & - & 98.78\% (\textit{30}) & - \\
\cite{jin2018hybrid} & Hybrid direct training & 99.49\% (\textit{400}) & - & - & 98.88\% (\textit{500}) & - \\
\cite{rathi2020enabling} & Hybrid direct training & - & 92.22\%  (\textit{250}) & 67.87\% (\textit{125}) & - & - \\
\cite{zhang2019spike} & Direct training & 99.62\% (\textit{400}) & - & - & - & - \\
\cite{wu2019direct} & Direct training & - & 90.53\% (\textit{12}) & - & 99.53\% (\textit{8}) & 60.50\% (\textit{8}) \\
\cite{zheng2021going} & Direct training & - & 93.16\% (\textit{6}) & - & - & 67.80\% (\textit{10}) \\
\cite{lee2020enabling} & Direct training & 99.59\% (\textit{50}) & 90.95\% (\textit{100}) & - & 99.09\% (\textit{100}) & - \\
\cite{cheng2020lisnn} & Direct training & 99.50\% (\textit{20}) & - & - & 99.45\% (\textit{20}) & - \\
\cite{fang2020exploiting} & Direct training & 99.46\% (\textit{25}) & - & - & 99.39\% (\textit{25}) & - \\
\cite{liu2020effective} & Direct training & - & - & - & 96.30\% (\textit{120}) & 32.2\% (\textit{80}) \\
\cite{fang2021deep} & Direct training & - & - & - & - & 70.2\% (\textit{8}) \\
\cite{fang2021incorporating} & Direct training & 99.72\% (\textit{8}) & 93.50\% (\textit{8}) & - & 99.61\% (\textit{10}) & 74.80\% (\textit{20}) \\
\cite{rathi2021diet} & DIET-SNN & - & 93.44\% (\textit{10}) & 69.67\% (\textit{5}) & - & - \\
\cite{li2021differentiable} & Dspike & - & 94.25\% (6) & 73.49\% (\textit{4}) & - & 75.40\% (\textit{10}) \\
\cite{guo2022recdis} & Recdis-SNN & - & \textbf{95.55}\% (\textit{6}) & 74.10\% (\textit{4}) & - & 72.42\% (\textit{10}) \\
\cite{deng2022temporal} & TET & - & 94.50\% (\textit{6}) & 74.62\% (\textit{4}) & - & 77.33\% (\textit{10}) \\
\cite{wangltmd} & Direct training & 99.60\% (\textit{4}) & 94.19\% (\textit{4}) & - & 99.65\% (\textit{15}) & 73.30\% (\textit{7}) \\ \hline
PSG\&PDA & Direct training & \textbf{99.72\%} (\textit{4}) & 95.35\% (\textit{6}) & \textbf{75.98\%} (\textit{4}) & \textbf{99.68\%} (\textit{15}) & \textbf{78.30\%} (\textit{7}) \\ \hline
\end{tabular}
\end{small}}
\vskip -0.1in
\end{table*}

\section{Ablation Study}
\subsection{Effect of Proposed Methods}
\begin{figure}
% \vskip 0.1in
  \centering
     \begin{subfigure}{0.49\columnwidth}
         \centering
         \includegraphics[width=\columnwidth]{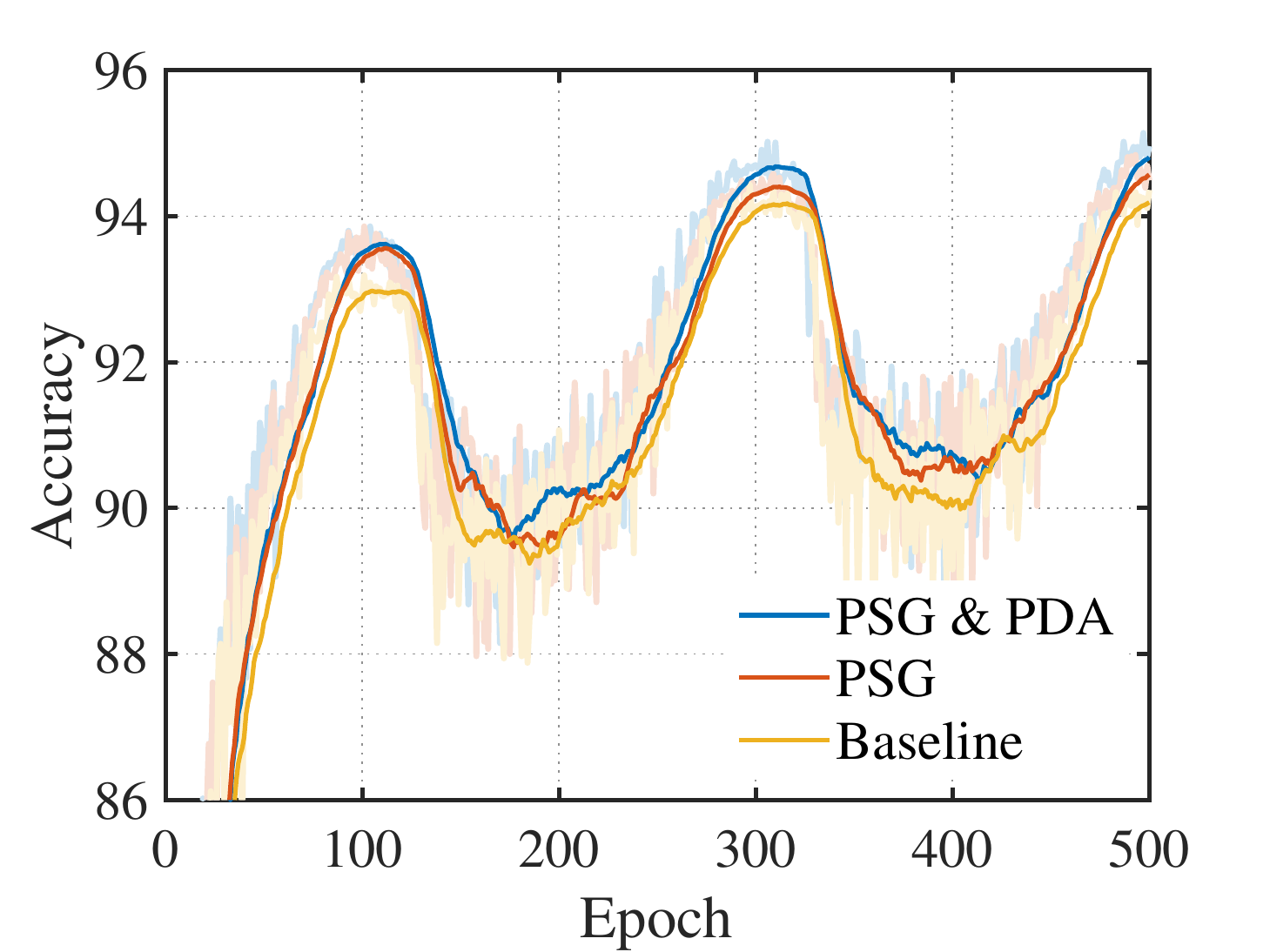}
         \caption{CIFAR10 \& $\gamma=1$.}
         \label{cifar10_gamma1}
     \end{subfigure}
     \hfill
     \begin{subfigure}{0.49\columnwidth}
         \centering
         \includegraphics[width=\columnwidth]{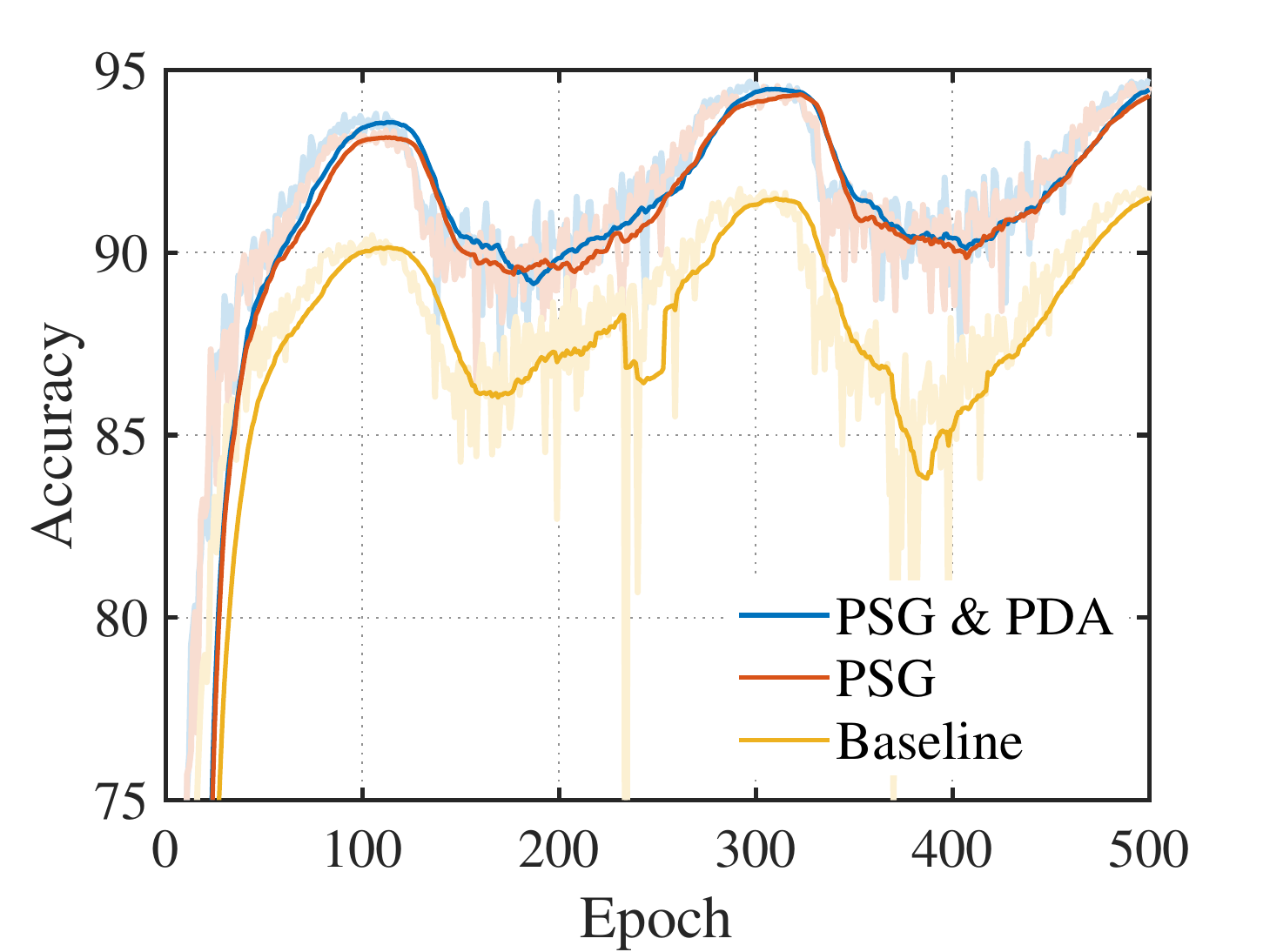}
         \caption{CIFAR10 \& $\gamma=2$.}
         \label{cifar10_gamma2}
     \end{subfigure}
     \hfill
     \begin{subfigure}{0.49\columnwidth}
         \centering
         \includegraphics[width=\columnwidth]{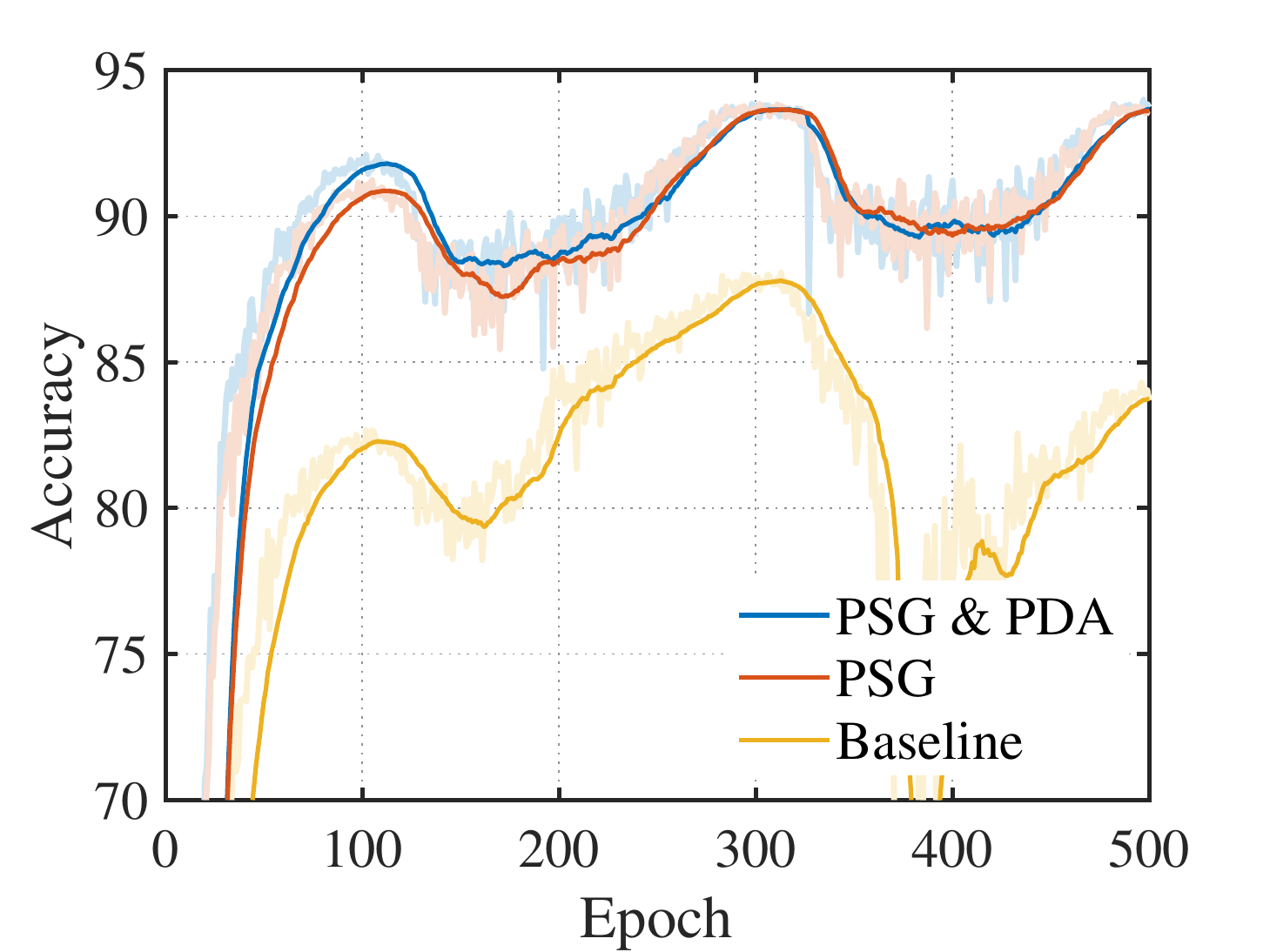}
         \caption{CIFAR10 \& $\gamma=5$.}
         \label{cifar10_gamma5}
     \end{subfigure}
     \hfill
     \begin{subfigure}{0.49\columnwidth}
         \centering
         \includegraphics[width=\columnwidth]{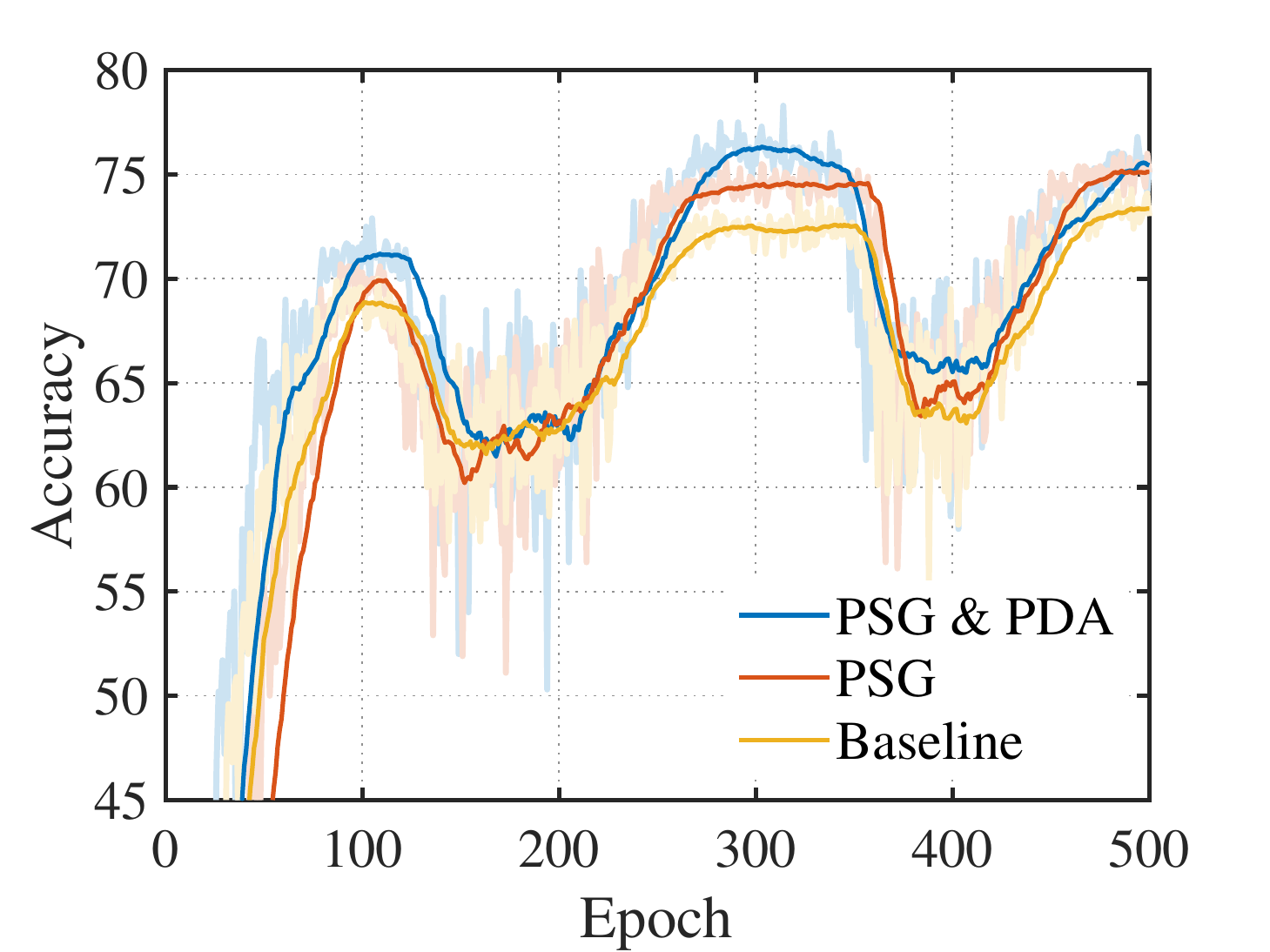}
         \caption{DVS-CIFAR10 \& $\gamma=1$.}
         \label{dvscifar10_gamma1}
     \end{subfigure}
     \hfill
     \begin{subfigure}{0.49\columnwidth}
         \centering
         \includegraphics[width=\columnwidth]{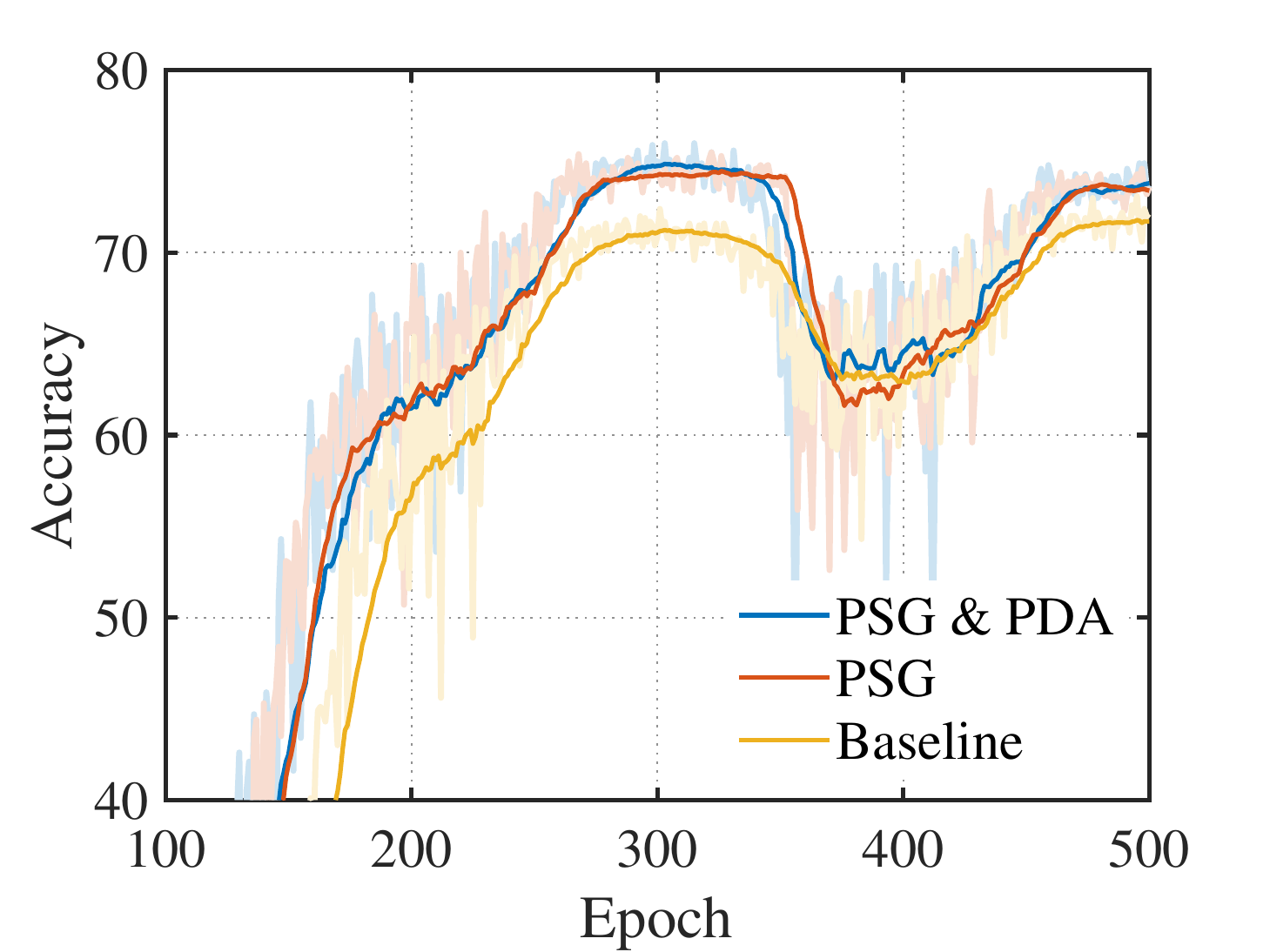}
         \caption{DVS-CIFAR10 \& $\gamma=2$.}
         \label{dvscifar10_gamma2}
     \end{subfigure}
     \hfill
     \begin{subfigure}{0.49\columnwidth}
         \centering
         \includegraphics[width=\columnwidth]{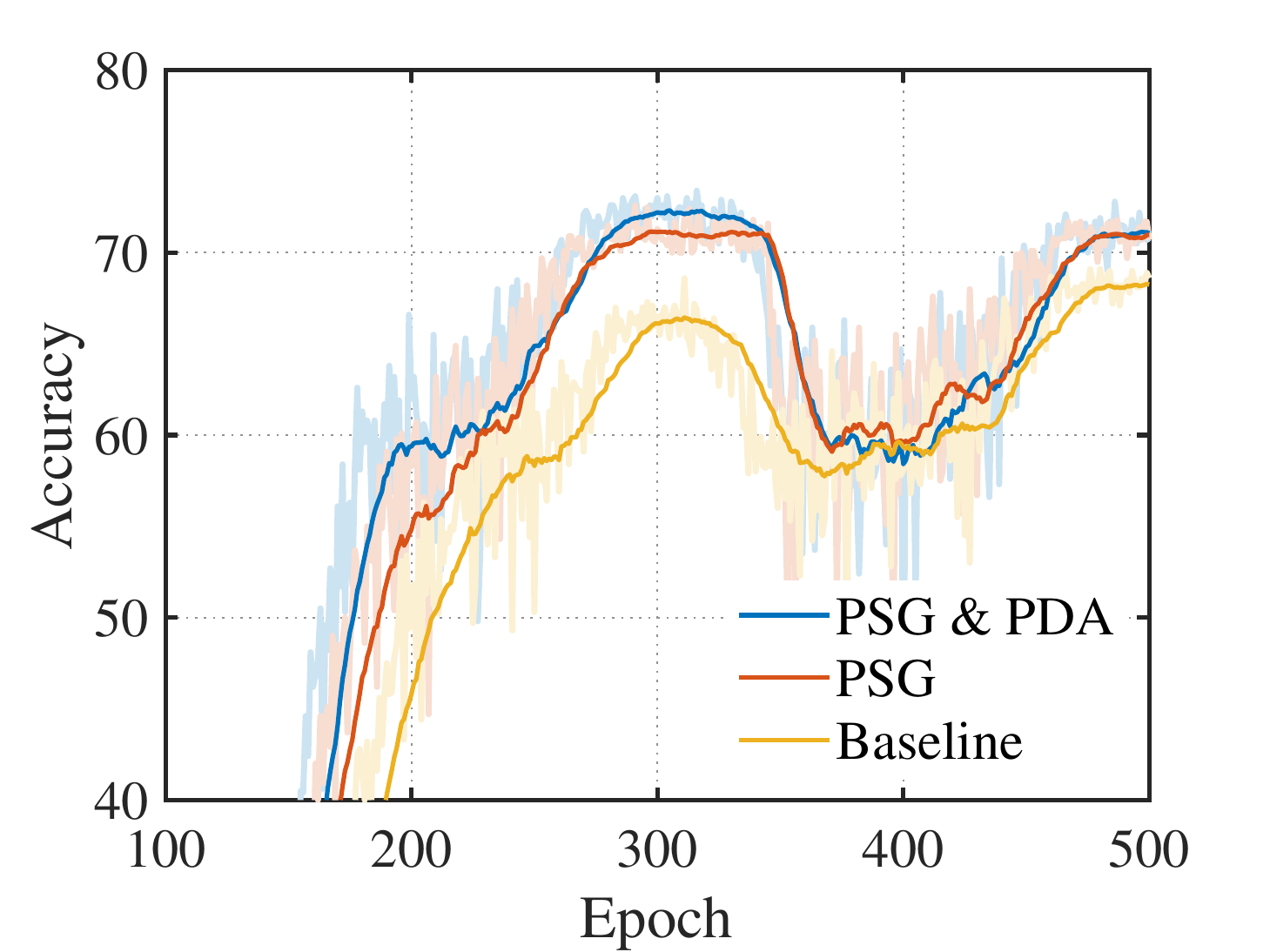}
         \caption{DVS-CIFAR10 \& $\gamma=5$.}
         \label{dvscifar10_gamma5}
     \end{subfigure}
\caption{Inference accuracies of SNNs on CIFAR10 ($T=2$) and DVS-CIFAR10 ($T=7$) with proposed methods implemented. Moving averages of 20 epoches are used, the original data is reflected as light-colour curves.}
\label{method_effect}
\vskip -0.1in
\end{figure}

To evaluate the effects of the PSG and PDA, a set of experiments have been conducted. We firstly train the plain SNNs followed by integrating the proposed PSG and PDA methods. Then, we use the modulated SNNs to do inference on test datasets to verify the impact of the proposed methods. The evaluation experiments were conducted on both static and neuromorphic datasets to obtain general observations. Based on observations from Fig.~\ref{method_effect} and Tab.~\ref{method_result}, we can conclude that constructed SNNs’ inference accuracy and convergence speed are consistently enhanced after PSG embedding, and are further improved with PDA, especially when the initial surrogate gradient is not properly set.

\begin{table}
\vskip -0.15in
\centering
\caption{Inference results of different datasets with the proposed methods implemented.}\label{method_result}
\vskip 0.1in
\resizebox{\columnwidth}{!}{
\begin{small}
\begin{tabular}{ccccc}
\hline
\multicolumn{1}{c}{\textbf{Dataset}} & \multicolumn{1}{c}{\textbf{Initial $\gamma$}} & \multicolumn{1}{c}{\textbf{Baseline}} & \multicolumn{1}{c}{\textbf{PSG}} & \multicolumn{1}{c}{\textbf{PSG\&PDA}} \\ \hline
\multirow{3}{*}{\begin{tabular}{@{}c@{}}CIFAR10 \\ ($T=4$)\end{tabular}}     & 1  & 94.78\%  & 95.00\% & 95.09\%    \\
                                                                             & 2  & 93.97\%  & 94.72\% & 94.86\%    \\
                                                                             & 5  & 89.35\%  & 94.14\% & 94.27\%    \\ \hline
\multirow{3}{*}{\begin{tabular}{@{}c@{}}DVS-CIFAR10 \\ ($T=7$)\end{tabular}} & 1  & 74.30\%  & 76.00\% & 78.30\%    \\
                                                                             & 2  & 73.10\%  & 75.50\% & 76.00\%    \\
                                                                             & 5  & 69.30\%  & 72.60\% & 73.40\%    \\ \hline
\end{tabular}
\end{small}}
\vskip -0.1in
\end{table}

\subsection{Robustness to Initial Surrogate Gradients}
\begin{figure}
\vskip -0.1in
  \centering
     \begin{subfigure}{0.49\columnwidth}
         \centering
         \includegraphics[width=\columnwidth]{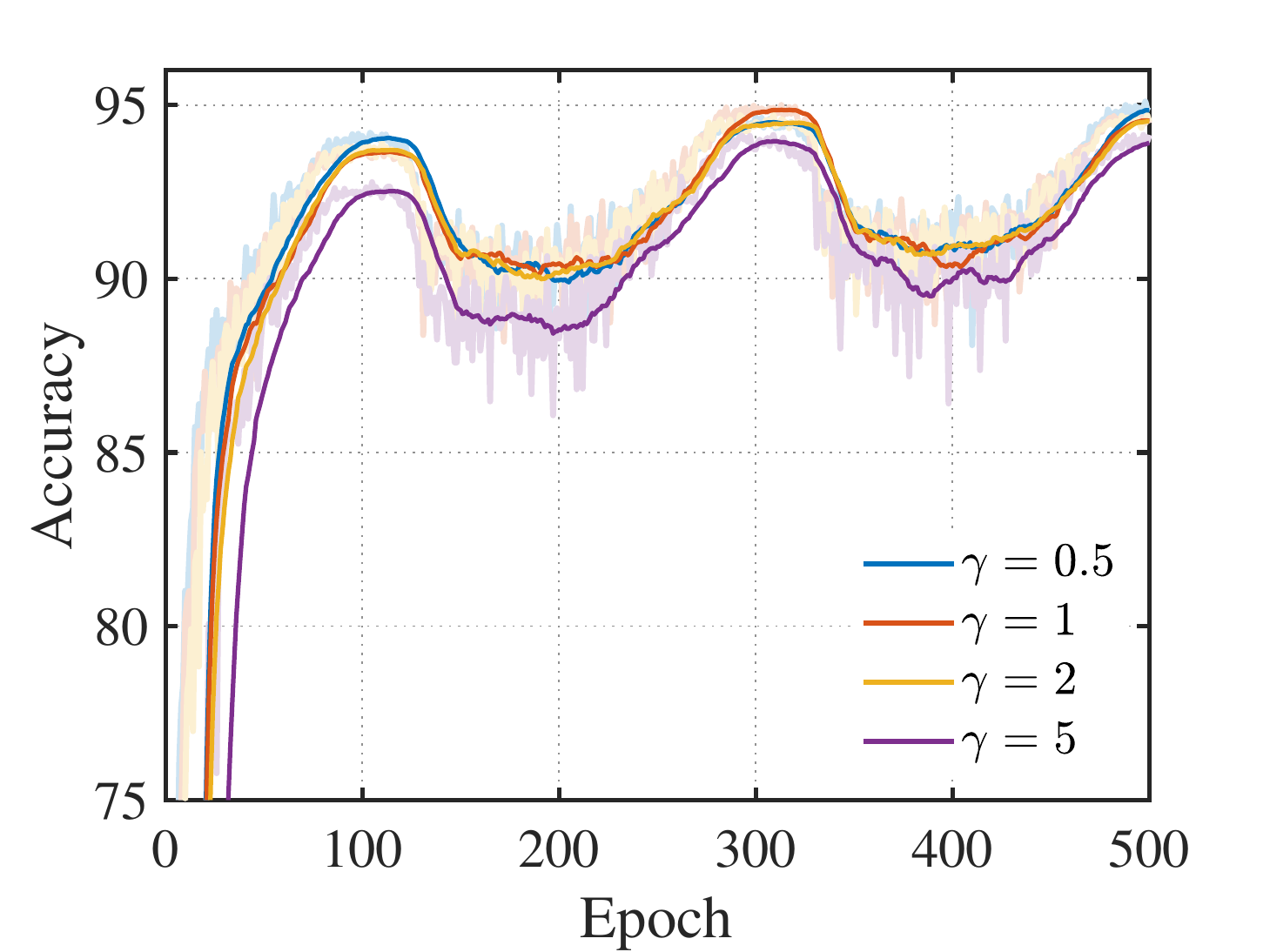}
         \caption{CIFAR10 ($T=4$).}
         \label{cifar10_diff_gamma}
     \end{subfigure}
     \hfill
     \begin{subfigure}{0.49\columnwidth}
         \centering
         \includegraphics[width=\columnwidth]{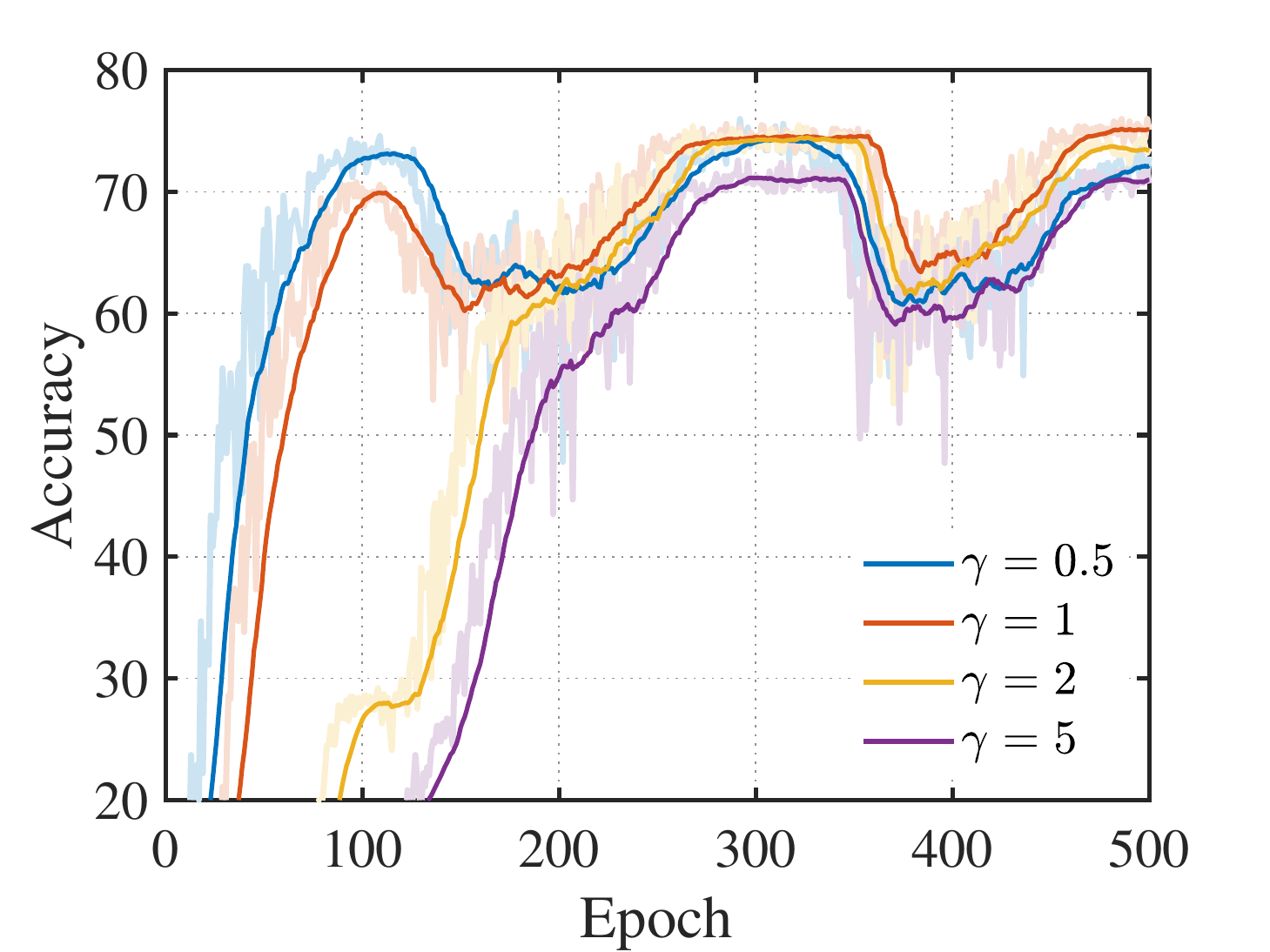}
         \caption{DVS-CIFAR10 ($T=7$).}
         \label{dvscifar10_diff_gamma}
     \end{subfigure}
\caption{Inference results of modulated SNNs with different initial surrogate gradient parameter values.}
\label{diff_gamma}
\vskip -0.1in
\end{figure}

\begin{figure}[ht]
\vskip -0.1in
  \centering
  \includegraphics[width=0.7\columnwidth]{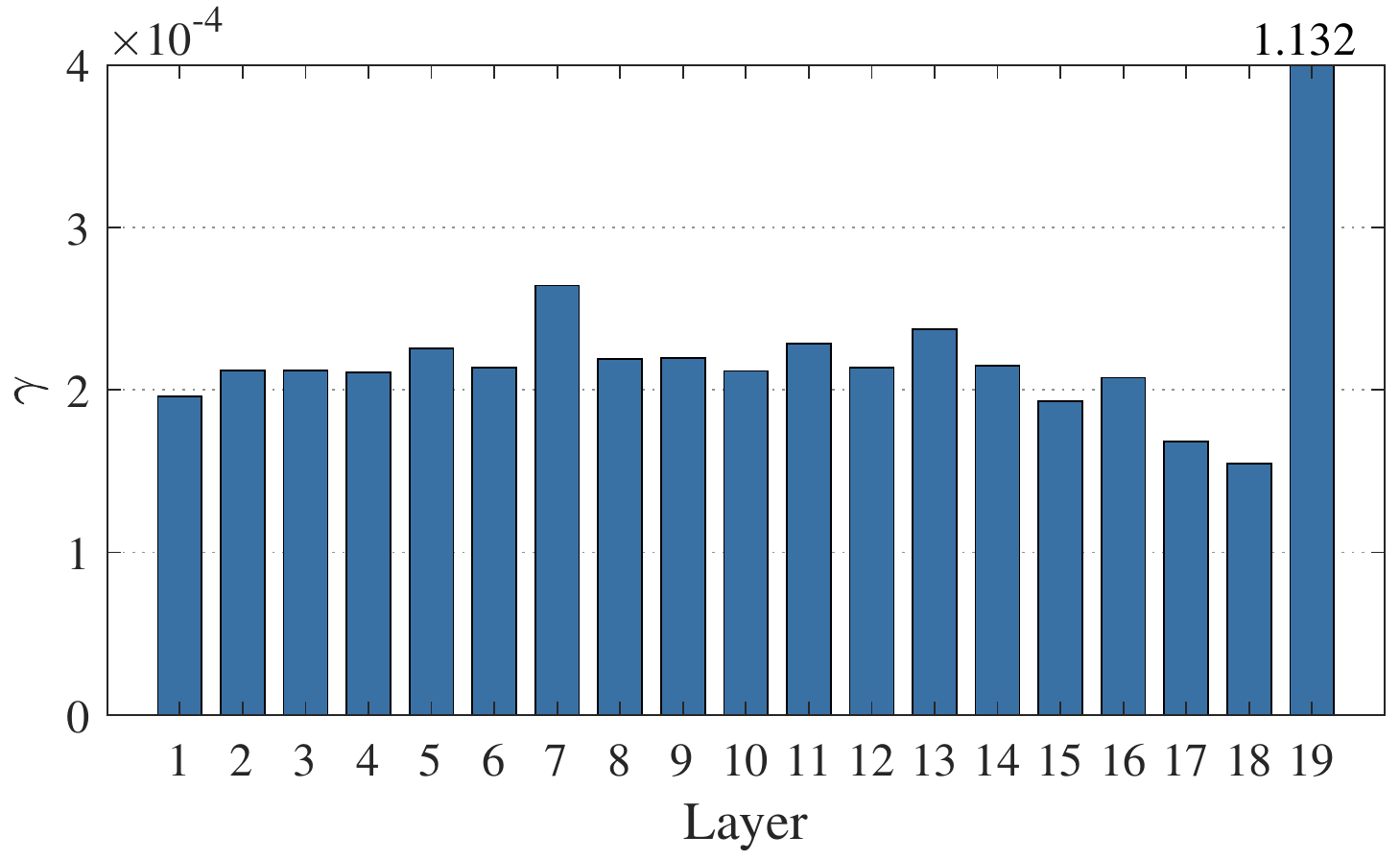}
\caption{Layer-wise surrogate parameter $\gamma$ values after training with PSG implemented.}
\label{layerwise_gamma}
% \vskip -0.1in
\end{figure}

We built several SNNs with the proposed methods and trained them with different initial surrogate gradient parameter values and plot their inference results, as shown in Fig.~\ref{diff_gamma}. It can be observed from the figure that SNNs’ performances tend to converge after training even if their initial surrogate gradients are significantly varied. Layer-wise surrogate gradient parameter values after training with PSG implemented are illustrated in Fig.~\ref{layerwise_gamma}, where it can be observed that shallow layers use small $\gamma$ values to ensure sufficiently wide neuronal update range while the final layer uses a much larger $\gamma$ to ensure proper network output. These results give a clear indication of the much higher robustness of our proposed SNNs after implementing PSG compared to plain SNNs shown in Fig.~\ref{different_SG_result}. Consequently, suboptimal surrogate gradient parameter selection will no longer be an issue after adopting the proposed PSG method.

\subsection{Computational Load}
\begin{figure}
% \vskip -0.1in
  \centering
  \includegraphics[width=0.7\columnwidth]{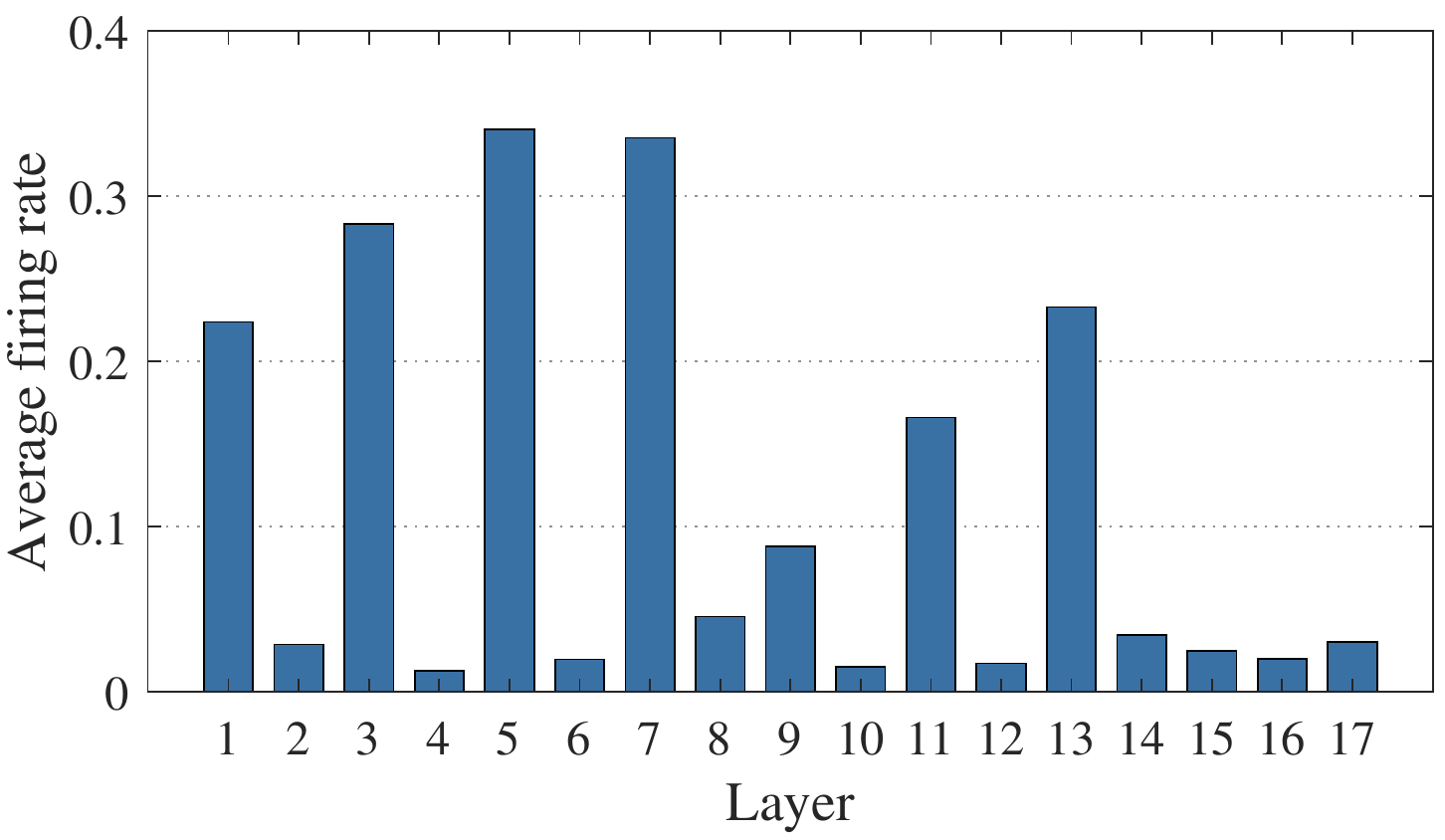}
\caption{Average firing rate of neurons in each layer of ResNet-19 SNN (fully-connected layers excluded).}
\label{layerwise_firing_rate}
\vskip -0.1in
\end{figure}

\begin{table*}[t]
\vskip -0.1in
\centering
\caption{The computational load and learnable parameters of SNNs. Changing percentage based on standard SNN is shown in brackets.}\label{comp_load}
\vskip 0.1in
\begin{small}
\begin{tabular}{ccccc}
\hline
\multicolumn{1}{c}{\textbf{Computational load}} & \multicolumn{1}{c}{\textbf{Standard SNN}} & \multicolumn{1}{c}{\textbf{PSG}} & \multicolumn{1}{c}{\textbf{PSG\&PDA}} & \multicolumn{1}{c}{\textbf{ANN}} \\ \hline
Training time per epoch & 233.36s   & 241.73s (+3.59\%)      & 355.60s (+52.38\%)     &         \\
% 233.35573s; 241.73432s; 355.59551s
Learnable parameters  & 12.63M      & 12.63M (+0.00015\%)    & 12.63M (+0.00015\%)    &         \\
\#Additions           & 2055.51M   & 1644.09M (-20.02\%)    & 1698.94M (-17.35\%)    & 2222.95M        \\
% standard firing rate: 0.2283396525; PSG firing rate: 0.1819008892 ; PSG&PDA firing rate: 0.1880923670
\#Multiplications     & 8.15M       & 8.15M (+0\%)           & 8.15M (+0\%)           & 2222.95M        \\ \hline
\end{tabular}
\end{small}
\vskip -0.1in
\end{table*}
To illustrate the computational load of our SNNs, we analyzed the training time, number of additions, number of multiplications, and number of learnable parameters of the network model after implementing the proposed methods on CIFAR10 dataset when $T=4$ in Tab.~\ref{comp_load}. Furthermore, the layer-wise neurons’ average firing rate is evaluated and presented as Fig.~\ref{layerwise_firing_rate}. In this work, ResNet-19 architecture is mainly used, and we also provide corresponding ANN statistics in the table for comparison. In SNNs, neurons’ states are binarized and only addition operations are needed when spiking neuron fires. Therefore, we use $r\times T\times A$ to calculate the number of additions in our SNN, where $r$ is the average firing rate, $T$ is the simulation timesteps, and $A$ is the addition count in corresponding ANN. It should be noted that multiplications are still needed in our model because real-value input is directly fed into the encoding layer and potentials instead of spikes are used as output in fully-connected layers to provide network predictions. It can be observed from Tab.~\ref{comp_load} that the proposed SNN significantly reduces computational operations by using sparse events compared to ANNs. As such, PDA and PSG will effectively enhance the SNN performance even with less computational load required due to more proper neuronal update triggered by the optimal surrogate gradients.

\subsection{Robustness to Neuromorphic Hardware Noise}
\begin{figure}
\vskip -0.1in
  \centering
     \begin{subfigure}{0.49\columnwidth}
         \centering
         \includegraphics[width=\columnwidth]{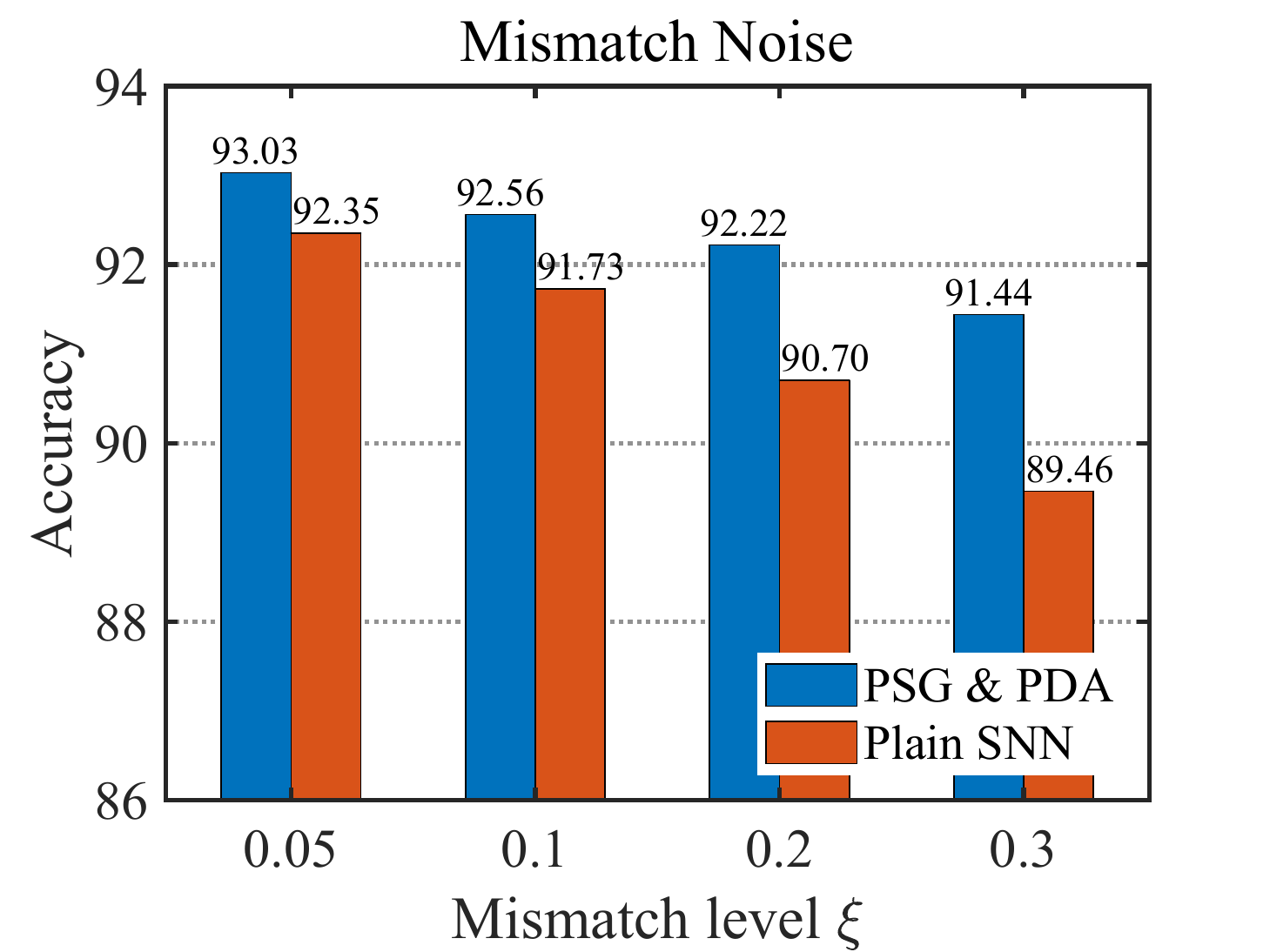}
     \end{subfigure}
     \hfill
     \begin{subfigure}{0.49\columnwidth}
         \centering
         \includegraphics[width=\columnwidth]{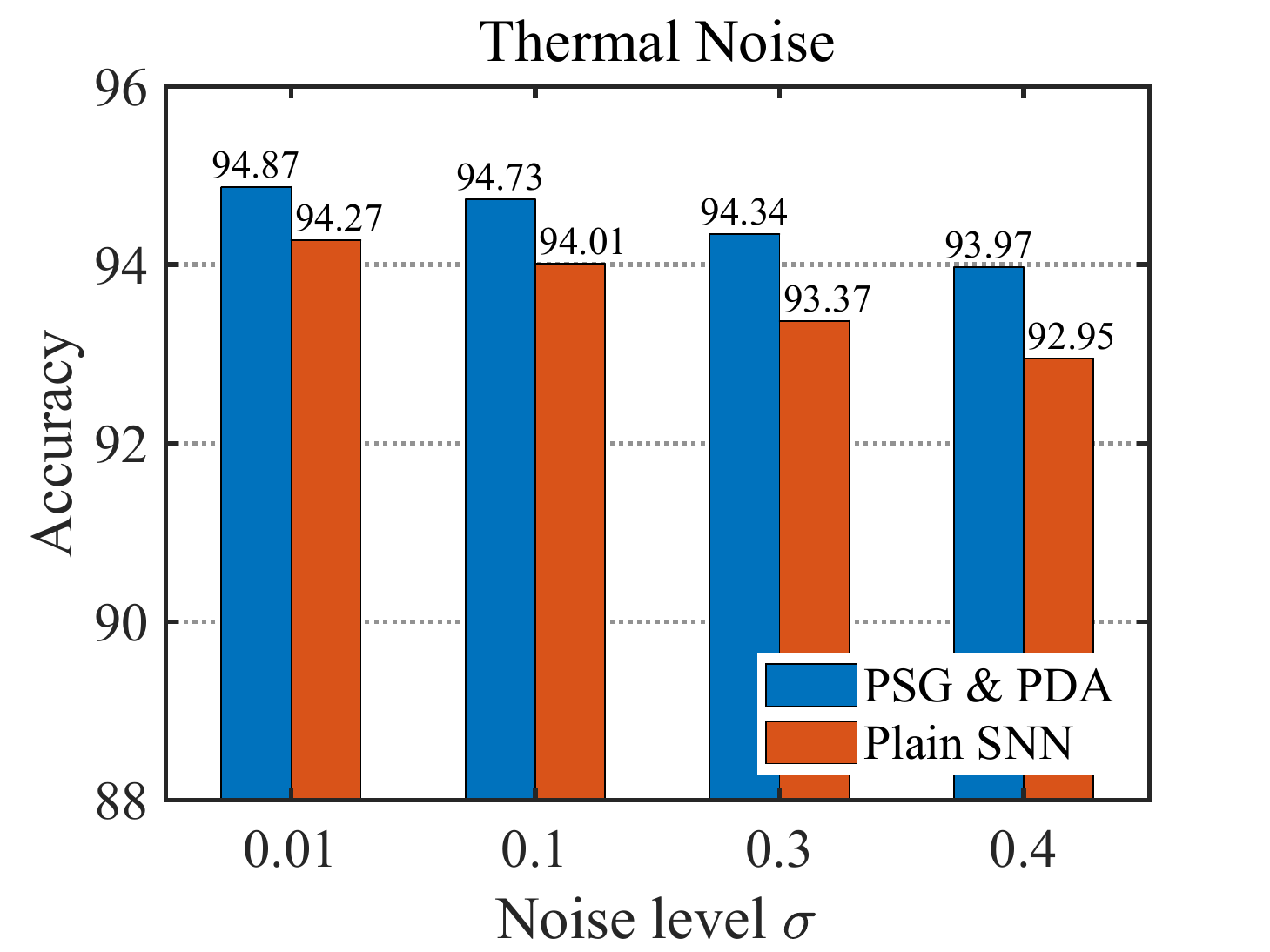}
     \end{subfigure}
\caption{Classification accuracies under different neuromorphic hardware-related noise.}
\label{hardware_noise}
\vskip -0.1in
\end{figure}

Mixed-signal neuromorphic hardware inevitably encounters various types of noise when deploying SNNs to perform machine learning inference tasks \cite{buchel2021supervised}. We verify our SNNs’ robustness on the two major neuromorphic hardware-related noise models: mismatch and thermal noise. When running on neuromorphic chips, network parameters such as membrane decay factor and neuronal thresholds may suffer from discrepancy due to chip process variation, resulting in mismatch noise in between neurons and synapses. In our experiment, simulated mismatch noise is introduced by incorporating Gaussian noise $\Psi'\leftarrow \mathcal{N}(\Psi,\xi(\Psi))$ on all mismatch parameters $\Psi$, where $\xi$ is the mismatch level. In addition, neuromorphic hardware always falls into thermal noise, represented as the unexpected noise on neuronal input. In Fig.~\ref{hardware_noise}, we compare the performance of SNNs with PDA and PSG implemented for the plain SNNs on CIFAR10 dataset using 2 timesteps for different noise levels. By evaluating model robustness, the proposed PDA and PSG methods demonstrate the capability of mitigating degradation triggered by neuromorphic hardware-related noise. 

\section{Conclusion and Discussion}
In this work, we propose PDA and PSG to redistribute neurons’ potentials before updating neuronal states and to find the optimal SG for each layer in the SNN. PDA can effectively normalize the neurons’ pre-activations and curb potential shift by minimizing the undesired potential distribution penalty on the proposed loss function. In PSG, by incorporating the procedure of optimizing SG into BPTT, SG is endowed with the capability of approaching its optimal expression automatically as network converges, such that degeneration caused by improper SG selection can be mitigated. To the best of our knowledge, PSG is the first method that focuses on rectifying SG automatically with BPTT in SNNs. In our SNNs, the first layer directly encodes input values into sparse spikes. As such, our SNNs can operate with high activation sparsity and low inference latency. We utilized different auxiliary functions for different types of layers to adapt distinctive neuronal states. Furthermore, we evaluate the computational load triggered by our methods, indicating that the designed PDA and PSG can greatly improve SNNs performance with even less computations incurred. It is also important to note that, our methods exhibit high robustness and remain operable under diverse neuromorphic hardware-related noise conditions, which effectively extends their application domains and more amenable for implementation on neuromorphic chips. Our SNNs report better or comparable accuracies using fewer timesteps on all datasets, demonstrating the superiority of the proposed strategies.

% Acknowledgements should only appear in the accepted version.
% \section*{Acknowledgements}

% \textbf{Do not} include acknowledgements in the initial version of
% the paper submitted for blind review.

% If a paper is accepted, the final camera-ready version can (and
% probably should) include acknowledgements. In this case, please
% place such acknowledgements in an unnumbered section at the
% end of the paper. Typically, this will include thanks to reviewers
% who gave useful comments, to colleagues who contributed to the ideas,
% and to funding agencies and corporate sponsors that provided financial
% support.

% In the unusual situation where you want a paper to appear in the
% references without citing it in the main text, use \nocite
% \nocite{langley00}

\bibliography{example_paper}
\bibliographystyle{icml2023}

\end{document}